\documentclass[3p,twocolumn]{elsarticle}
\usepackage{hyperref}
\journal{Journal of \LaTeX\ Templates}
\usepackage{amssymb}
\usepackage{graphicx}
\usepackage{textcomp}
\usepackage{amsmath,amssymb,amsfonts}
\usepackage{algorithmic}
\usepackage{graphicx}
\usepackage{caption}
\usepackage{textcomp}
\usepackage{color}
\usepackage[dvipsnames]{xcolor}
\usepackage{subfigure}
\usepackage{multirow}
\usepackage{rotating}
\usepackage{diagbox}
\usepackage{stfloats}
\usepackage{float}
\usepackage{bm}
\usepackage{booktabs}
\usepackage{float}
\usepackage{multirow}
\usepackage{soul}
\soulregister\cite7
\soulregister\citep7
\soulregister\citet7
\soulregister\ref7 
\soulregister\pageref7 
\usepackage{appendix}
\usepackage{gensymb}
\usepackage{float} 

\begin{document}

\begin{frontmatter}

\title{TOD-CNN: An Effective Convolutional Neural Network for 
Tiny Object Detection in Sperm Videos}

\author[myaddress]{Shuojia Zou}

\author[myaddress]{Chen Li\corref{mycorrespondingauthor}}
\cortext[mycorrespondingauthor]{Corresponding author}

\ead{lichen201096@hotmail.com}
\author[1address]{Hongzan Sun}
\author[2address]{Peng Xu}
\author[myaddress]{Jiawei Zhang}
\author[myaddress]{Pingli Ma}
\author[3address]{Yudong Yao}
\author[4address]{Xinyu Huang}
\author[4address]{Marcin Grzegorzek}

\address[myaddress]{Microscopic Image and Medical Image Analysis Group, College of 
Medicine and Biological Information Engineering, Northeastern University, Shenyang, China} 
\address[1address]{Shengjing Hospital, China Medical University, Shenyang, China} 
\address[2address]{Jinghua Hospital, Shenyang, China}
\address[3address]{Department of Electrical and Computer Engineering, 
Stevens Institute of Technology, US}
\address[4address]{Institute of Medical Informatics, University of Luebeck, Luebeck, Germany}

\begin{abstract}
The detection of tiny objects in microscopic videos is a problematic point, especially 
in large-scale experiments. For tiny objects (such as sperms) in microscopic videos, 
current detection methods face challenges in fuzzy, irregular, and precise positioning 
of objects. In contrast, we present a convolutional neural network for tiny object 
detection (TOD-CNN) with an underlying data set of high-quality sperm microscopic videos 
(111 videos, $>$ 278,000 annotated objects), and a graphical user interface (GUI) is designed to employ and test the proposed model effectively. TOD-CNN is highly accurate, achieving $85.60\%$ 
AP$_{50}$ in the task of real-time sperm detection in microscopic videos. To demonstrate the importance of sperm 
detection technology in sperm quality analysis, we carry out relevant sperm quality evaluation metrics and compare them with the diagnosis results from medical doctors.
\end{abstract}


\end{frontmatter}

\section{Introduction\label{sec:Introduction}}

Sperm is necessary for the human and mammal reproductive process, 
which plays an important role in human reproduction and animal breeding~\cite{gadadhar2021TGCA}. 
With the continuous development of computer technology, researchers have tried to use computer-aided  image analysis in many fields, 
such as whole-slide image analysis~\cite{Li2022ACRO}, histopathology image analysis~\cite{li2022AHCR, li2022ACROM, li2020AROCH}, cytopathological analysis~\cite{RAHAMAN2021DADLF, Rahaman2020ASFCCI}, COVID-19 image analysis~\cite{rahaman2020IOCSF, li2020ASMID}, and microorganism counting~\cite{zhang2021ACROI}. In addition, in the field of semen analysis and diagnosis, researchers have also proposed many \emph{Computer Aided Semen Analysis} (CASA) systems~\cite{zhao2022ASOS}.
As the first step of the CASA system, 
sperm detection is one of the most important parts to support the reliability of sperm analysis results~\cite{zhao2020ASOSD}.
At present, most sperm detection techniques~\cite{elsayed2015DOC,urbano2016ATA, li2020FFF, yang2014HTA} are 
based on traditional image processing techniques such as thresholding, edge detection 
and contour fitting. However, for many techniques, the detection results require 
manual intervention. The common difficulties for sperm detection mainly include the small size, uncertain morphologies and low contrast of the sperms, which are difficult for locating. Moreover, there are lots of similar impurities in the samples for misleading (as shown in Section 4 Fig.~\ref{fig1}).

In recent years, more and more excellent object detection models are constantly proposed~\cite{zou2019ODI}, such as \emph{Region-based CNN} (RCNN) series 
models~\cite{girshick2014RFHF, girshick2015FRCNN, ren2016FRTR, he2017mask}, 
\emph{You Only Look Once} (YOLO) series models~\cite{redmon2016YOLO, redmon2017yolo9000, redmon2018yolov3, bochkovskiy2020yolov4}, \emph{Single Shot Multibox Detector} 
(SSD)~\cite{liu2016ssd}, and RetinaNet~\cite{lin2017FLFD}.
The performance of \emph{Convolutional Neural Networks} (CNN) has obviously surpassed the complex 
classic image processing algorithms in the field of medical image processing~\cite{gu2020AARO, wang2022EE-Net, yang2017ILCI}, 
which makes it possible to use deep learning methods to perform real-time sperm object detection tasks in sperm microscopic videos. However, the accuracy of sperm object detection 
is still lower than that of object detection under conventional scales~\cite{li2017PGA}. 
Hence, techniques such as feature fusion and residual networks are used in our method 
to improve the detection performance in this field. The technologies above are applied 
to build an easy-to-operate sperm detection model (TOD-CNN), and an  AP$_{50}$ of 
$85.60\%$ is achieved in the task of sperm detection for microscopic videos.

The workflow of the proposed TOD-CNN detection method is summarized as follows (as shown in Section 3 Fig.~\ref{fig2}): 
(a) Training and Validation Data: The training and validation data contains 80 sperm microscopic videos and corresponding annotation data with 
the location and category information of sperms and impurities. (b) Data Preprocessing: 
The sperm microscopic video is divided into frames to obtain one by one sperm microscopic images, and the object information is annotated by using LabelImg software. (c) 
Training Process: The TOD-CNN model is trained and the best model is saved to perform 
sperm object detection. (d) Test Data: The test data contains 21 sperm microscopic videos.

The main contributions of this paper are as follows:
\begin{itemize}
\item Build an easy-to-operate CNN for sperm detection, namely TOD-CNN (Convolutional 
Neural Network for tiny object detection).
\item TOD-CNN has excellent detection results and real-time detection ability in the task of tiny object detection in sperm microscopic video, achieving $85.60\%$ AP$_{50}$ and $35.7$ \emph{frames per second} (FPS).
\end{itemize}

The structure of this paper is as follows: Section 2 introduces the existing sperm object detection methods based on traditional methods, machine learning methods, and deep learning methods. Section 3 illustrates the detailed design of TOD-CNN.  Section 4 introduces the data set used in the experiment, experiment settings, evaluation methods, and results. Chapter 5 is conclusion.

\section{Related Work\label{sec:Related Work}}

\subsection{Existing Sperm Object Detection Methods}
\subsubsection{Traditional Methods}
Traditional methods mainly include three types, which are threshold-based methods, 
shape fitting methods, and filtering methods. Threshold-based methods: Urbano et al.~\cite{urbano2016ATA} use Gaussian filter to enhance the image, and then 
the image is binarized using the Otsu~\cite{otsu1979ATSM} threshold method, and the result is morphologically operated to determine the position of the sperm; Elseyed et al.~\cite{elsayed2015DOC} use several certain frames to generate the background information, 
then the background information is subtracted from the original image (to suppress noise). Finally, the Otsu threshold is applied to determine the position of the sperm. 
Shape fitting methods: Zhou et al.~\cite{zhou2009EMS} use a rectangular area which is similar to the shape of 
the object (sperm) to fit the object, and then the position of the sperm 
is described by the parameters of the rectangle. Yang et al.~\cite{yang2014HTA} use an ellipse to approximate the sperm head, and the improved multiple birth and 
cut algorithm based on marked point processes~\cite{soubies2013ASA} is used to detect 
and locate the head of the sperm through modelling. Filtering methods: 
Ravanfar et al.~\cite{ravanfar2011LCS} select several suitable structural elements firstly, and then 
the operation based on Top-hat is used to filter the image sequence to achieve the purpose of separating sperm and other debris.  
Nurhadiyatna et al.~\cite{nurhadiyatna2014CAIO} use the \emph{Gaussian Mixture Model} (GMM) enhanced by the Hole Filling Algorithm as the probability density function to predict the probability of each pixel in the 
image belongs to the foreground and the background. The researchers found that the 
calculation amount of this method is significantly less than other methods.

\subsubsection{Machine Learning Methods}
The unsupervised learning method is the most used machine learning method. 
Berezansky et al.~\cite{berezansky2007SATO} use the Spatio-Temporal Segmentation to detect sperm 
by segmentation, integrating $k$-means, GMM, mean shift, and other segmentation methods. 
Shi et al.~\cite{z2006RATA} use the optical capture method for sperm detection.

\subsection{Deep Learning Based Object Detection Methods}
Deep learning methods are widely used in many artificial intelligent fields, for example classification~\cite{wang2016self, yi2016improved, KOSOV2018EMCU, li2015AOCI}, segmentation~\cite{ZHANG2021107885, Kulwa2019ASSFMI, SUN2020GHIS} and object detection~\cite{cui2018detection, shen2015ITOIP}. Furthermore, some widely recognized general object detection models that have been proposed in recent years are introduced bellow.

\subsubsection{One-stage Object Detector}
The YOLO series models~\cite{redmon2016YOLO, redmon2017yolo9000, redmon2018yolov3, bochkovskiy2020yolov4}, RetinaNet~\cite{lin2017FLFD}, and SSD~\cite{liu2016ssd} are prominent representatives of one-stage object detectors. The one-stage object detectors are based on the idea of regression, which can directly output the final prediction results from the input images without generating suggested regions in advance. 
YOLO series models: They use Darknet as the backbone of the model to extract features 
from the image. The v1, v2, v3, v4 are successively proposed by improving the backbone 
network structure, improving the loss function, using batch normalization, feature 
pyramid network~\cite{lin2017FPN}, spatial pyramid pooling network~\cite{he2015SPPI}, 
and other optimization methods. RetinaNet: It uses ResNet~\cite{he2016DRLF} and the 
feature pyramid network as the backbone of the model, whose main contribution is 
proposing a focus loss function. The focus loss function solves the imbalance between the 
number of foreground and background categories in a single-stage object detector. 
SSD: It uses VGG16~\cite{simonyan2014VDCN} as the basic model and then adds a new 
convolutional layer based on VGG16 to obtain more feature maps for detection and 
generates the final prediction result by fusing the prediction results of 6 
feature maps.

\begin{figure*}[b]
\centering
\includegraphics[width=0.9\linewidth]{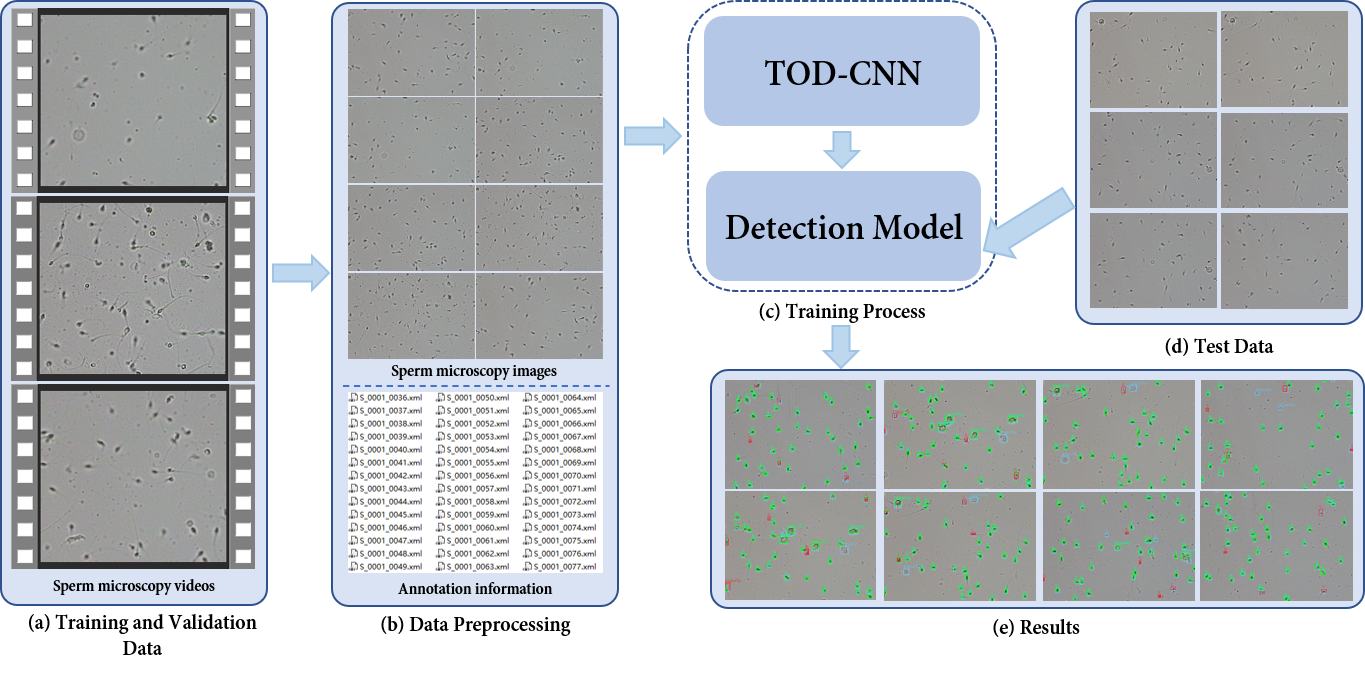}
\caption{The workflow of the proposed sperm object detection method using TOD-CNN.}
\label{fig2}
\end{figure*}

\subsubsection{Two-stage Object Detector}
The models based on R-CNN series are classical two-stage object detectors. The 
R-CNN~\cite{girshick2014RFHF} generates proposal regions through the selective search 
algorithm. Then the features of the proposal regions are extracted by using CNN. 
Finally, the SVM classifier is used to predict the objects in each region and identify 
the category of the objects. The Fast R-CNN~\cite{girshick2015FRCNN} no longer extracts 
features for each proposal region. The features of entire image is extracted using CNN, then each proposal region and corresponding features are mapped. Besides, the Fast R-CNN 
uses a multi-task loss function, allowing us to train the detector and bounding box 
regressor simultaneously. The Faster R-CNN~\cite{ren2016FRTR} replaces the selective search algorithm with Region Proposal Network, which can help CNN to generate proposal regions and detect objects simultaneously.

\section{TOD-CNN based Sperm Detection Method in Microscopic Image}
\label{s:methods}
Sperm detection is always the first step in a CASA system, which determines the 
reliability of the results of sperm microscopic video analysis. However, the 
existing algorithms cannot accurately detect sperms. Therefore, we follow the 
idea of YOLO~\cite{redmon2016YOLO, redmon2017yolo9000, redmon2018yolov3, bochkovskiy2020yolov4}, ResNet~\cite{he2016DRLF}, Inception-v3~\cite{szegedy2016RTIA}, 
and VGG16~\cite{simonyan2014VDCN} models and propose a novel one-stage deep 
learning based sperm object detection model (TOD-CNN). The workflow of the proposed TOD-CNN detection approach is shown in Fig.~\ref{fig2}.

\subsection{Basic Knowledge}
In this section, the methods related to our work are introduced, including YOLO, 
ResNet, Inception-v3, and VGG16 models.

\subsubsection{Basic Knowledge of YOLO}
YOLO series models solve the object detection task as a regression problem. The 
YOLO series models remove the step of generating the proposal region in the 
two-stage object detector and accelerate the detection process. 
YOLO-v3~\cite{redmon2018yolov3} is the most popular model in the YOLO series models 
due to its excellent detection performance and speed.

The YOLO-v3 model mainly consists of four parts, which are preprocessing, backbone, 
neck, and head. The preprocessing: The $k$-means algorithm is used to cluster nine 
anchor boxes in the data set before training. The backbone: YOLO-v3 uses Darknet53 
as the backbone network of the model. Darknet53 does not have maxpooling layers and 
fully connected layers. The fully convolutional network can change the size of the tensor by changing the strides of the convolutional kernel. In addition, 
Darknet53 follows the idea of ResNet and adds a residual module to the network to 
solve the vanishing gradient problem of the deep network. The neck: The neck of the YOLO-v3 
model draws on the idea of the feature pyramid network~\cite{lin2017FPN} to enrich 
the information of the feature map. The head: The head of YOLO-v3 outputs 3 feature maps with different sizes and then detects large, medium, and small size 
objects of the three feature maps with three sizes.

\subsubsection{Basic Knowledge of ResNet}
ResNet~\cite{he2016DRLF} is one of the most widely used feature extraction CNNs due 
to its practical and straightforward structure. With the continuous deepening of CNN, 
the model's performance cannot be continuously improved, and the accuracy may even 
decrease. However, ResNet proposes the Shortcut Connection structure to solve the problems above. The identity mapping operation and residual mapping operation are 
included in the Shortcut Connection structure. The identity mapping is to pass the 
current feature map backward through cross-layer transfer (when the dimension of 
feature map does not match, a $1 \times 1$ convolution operation is used to adjust 
the dimension of feature map).  Residual mapping is to pass the current feature map 
to the next layer after convolution operation. A Shortcut Connection structure 
contains one identity mapping operation and two or three residual mapping operations in general.

\subsubsection{Basic Knowledge of Inception-v3 and VGG16}
In Inception-v3~\cite{szegedy2016RTIA}, to reduce the parameters and ensure the performance of the model, an operation that replaces $N \times N$ convolution kernels with $1 
\times N$ and $N \times 1$ convolution kernels is proposed. The receptive fields of 
$1 \times N$ and $N \times 1$ convolution kernels and $N \times N$ convolution kernels 
are the same, where the former has less parameters than the latter. 
In addition, the 
Inception-v3 model can support multi-scale input, which can use convolution kernels with different sizes to perform convolution operations on the input images, and then 
the input feature maps can be connected to generate the final feature map.

VGG16~\cite{simonyan2014VDCN} model includes 13 convolutional layers, 3 fully connected layers, and 5 maxpooling layers. The most prominent feature of the VGG16 
model is its simple structure. All convolutional layers use the same convolution 
kernel parameters, and all pooling layers use the same pooling kernel parameters. 
Although the VGG16 model has a simple structure, it has strong feature extraction capabilities.

\subsection{The Structure of TOD-CNN}
\begin{figure*}[h]
\centering
\includegraphics[width=0.95\linewidth]{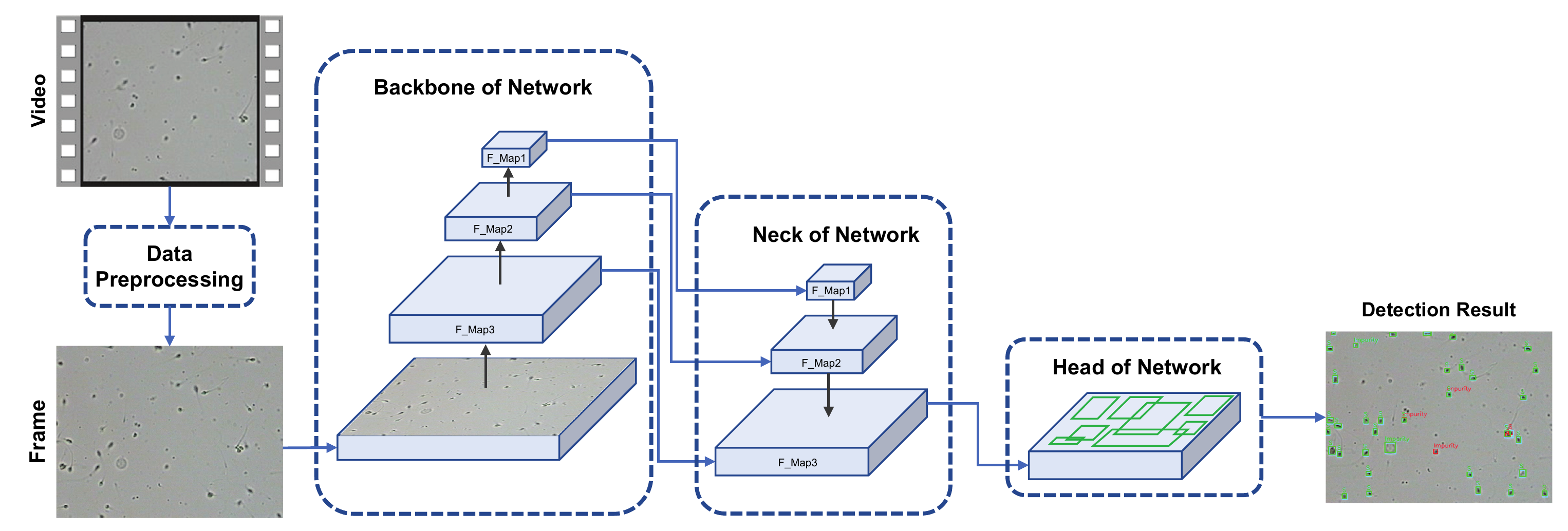}
\caption{The architecture of TOD-CNN.}
\label{fig3}
\end{figure*}

The TOD-CNN model, which refers to the YOLO series model, can regard the object 
detection task as a regression problem for fast and precise detection. The architecture of TOD-CNN is shown in Fig.~\ref{fig3}, where the entire network 
is composed of four parts: Data preprocessing, backbone of the network, neck of the 
network and head of the network. The detailed implementation of each part is introduced 
in detail below.

\subsubsection{Data Preprocessing}
The object detection task in the video is essentially based on image processing.
Therefore, it is necessary to split the sperm microscopic video into continuous
frames (single images). 
However, due to the movement of the lens during the sperm microscopic video shooting 
process, there are some blurred frames in the sperm microscopic video. 
After analysing the grayscale histogram of frames, there is an obvious difference 
between the grayscale distribution of the blurred frame and the normal frame. 
Therefore, the blurred frames can be solved by deleting images whose 
Otsu threshold is less than a certain threshold (from articial experience). 
In addition, TOD-CNN is an anchor-based object detection model. Therefore, the 
$k$-means algorithm is used to cluster a certain number (TOD-CNN uses six) of anchor boxes in data set to train the model.

\subsubsection{The Backbone of TOD-CNN}
\begin{figure*}[h]
\centering
\includegraphics[width=0.96\linewidth]{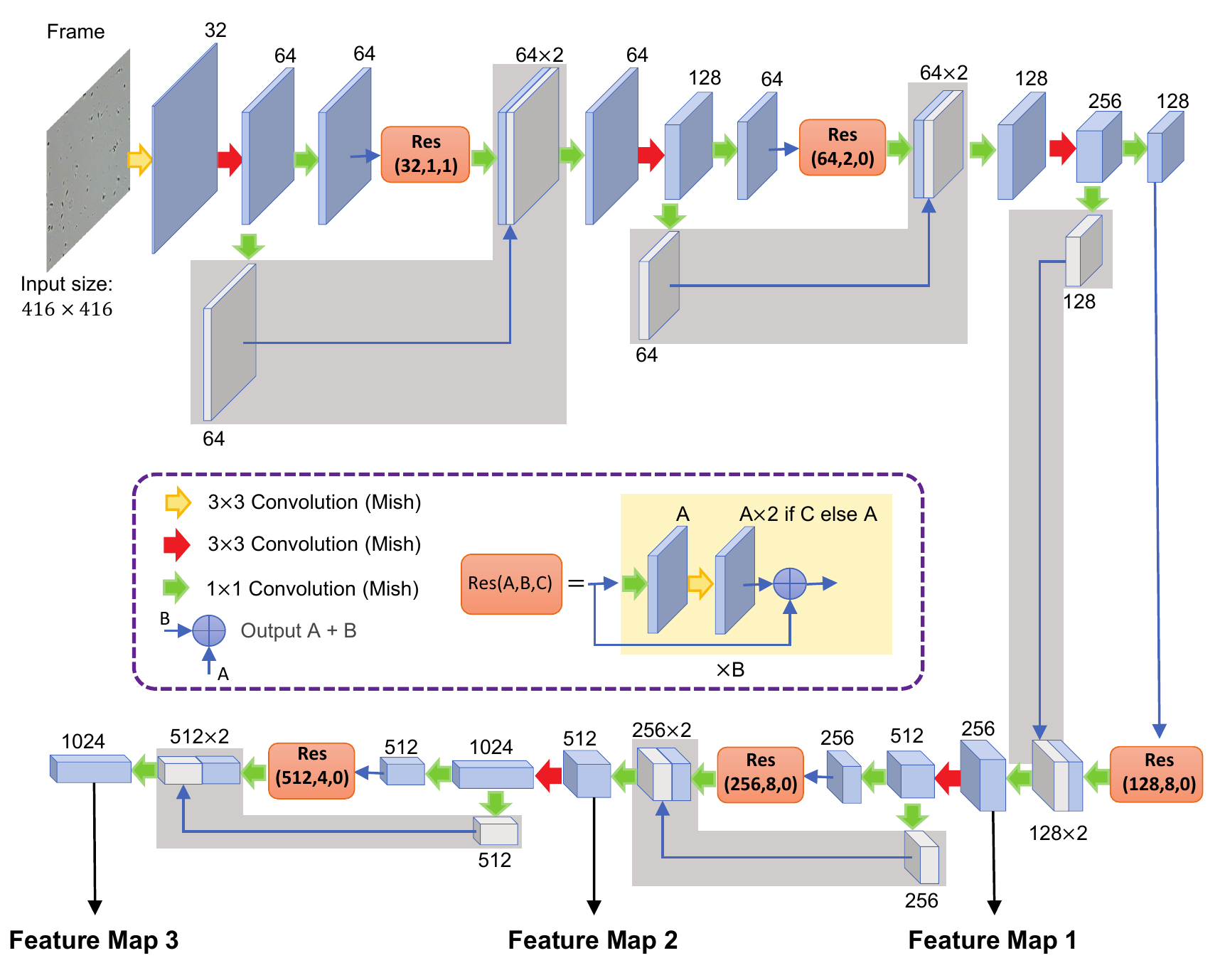}
\caption{The architecture of TOD-CNN backbone.}
\label{fig4}
\end{figure*}

A straight forward backbone structure with cross-layer concatenate operation is 
designed, which is shown in Fig.~\ref{fig4}. 
However, in a fully convolutional network, as the structures of CNNs continue to 
deepen, the semantic information of the feature map becomes more and more 
abundant, while the location information of the feature map constantly decreases. 
As a result, the network can improve the classification performance but may reduce the 
positioning accuracy. 
Our work focuses on detecting tiny objects and accurate locating, which needs to 
maintain precise local information. Therefore, we 
enhance the transfer of location information (transferring shallow features to 
deep layers) through the following methods: 
First, we refer to the residual idea of ResNet, the Shortcut Connection structure 
provides the approach for transferring local information with a cross-layer add 
operation, which is used in TOD-CNN (as shown in Res (A, B, C) in Fig.~\ref{fig4}); second,  based on the straightforward backbone 
structure, a cross-layer concatenate operation is applied to enhance the transfer of local information (as shown in grey shaded part in Fig.~\ref{fig4}).

The detailed design of TOD-CNN backbone is shown in Fig.~\ref{fig4}, where the input size of the backbone is $416 \times 416$, the yellow arrow indicates convolution operations with a kernel size of $3 \times 3$  and stride of 1 (each use filtering with padding and followed by a Mish activation), the red arrow indicates convolution operations with a kernel size of $3 \times 3$  and stride of 2 (each followed by a Mish activation), 
and the green arrow indicates convolution operations with a kernel size of $1 \times 1$  and stride of 1 (each use filtering with padding and followed by a Mish activation).

\begin{figure*}[htbp!]
\centering
\includegraphics[width=0.96\linewidth]{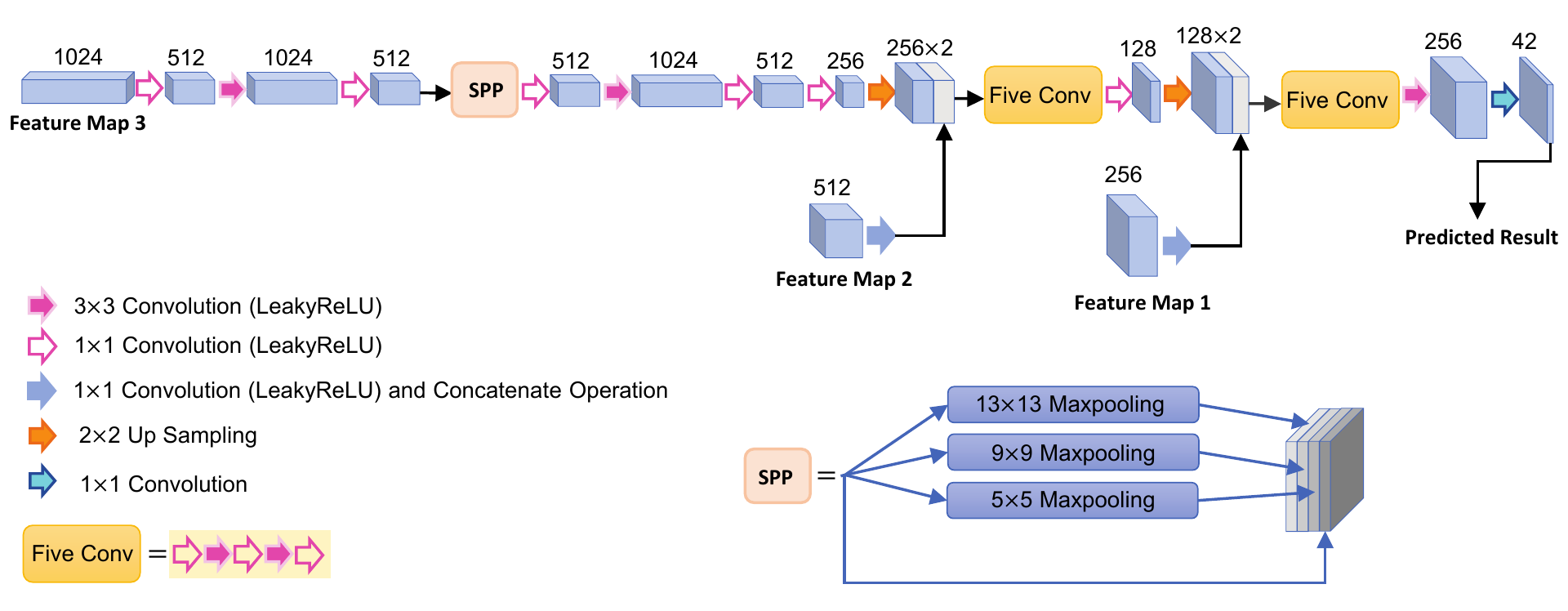}
\caption{The architecture of TOD-CNN neck.}
\label{fig5}
\end{figure*}

\subsubsection{The Neck of TOD-CNN}
In object detection model, the main purpose of model neck is to integrate feature 
informations extracted from model backbone. The neck structure of TOD-CNN is shown 
in Fig.~\ref{fig5}. In fact, there are abnormal morphological sperms (very big) 
and some other impurities (such as bacteria, protein lumps and bubbles) in semen. 
These sperms and impurities are significantly different from normal sperm in size. 
Therefore, to collect multi-scale information, we have adopted spatial pyramid 
pooling operation~\cite{he2015SPPI} to integrate multi-scale information into TOD-CNN neck. In addition, due to the small sizes of tiny objects, the information of tiny objects might be easily lost in down-sampling process. In order to solve this problem, 
the feature fusion method is used in TOD-CNN neck, where the shallow and deep feature 
maps are fused by upsampling to avoid the loss of tiny object information.

The detailed design of TOD-CNN neck are shown in Fig.~\ref{fig5}, where all convolution operations are with stride of 1 (each use filtering with padding ), and the detailed illustration of the kernel size and activation function is shown in Fig.~\ref{fig5}. Finally, TOD-CNN neck outputs a feature map of size $($input size$ /\ 8) \times ($input size$ /\ 8) \times 42$, where 42 is the number of anchor boxes$\ (6) \times 7$, because each anchor box needs to have 7 parameters: the relative center coordinates, the width and the high offset, class, and confidence, the details are explained in Section 3.2.4.

\subsubsection{The Head of TOD-CNN}

\begin{figure}[htbp!]
\centering
\includegraphics[scale=0.53]{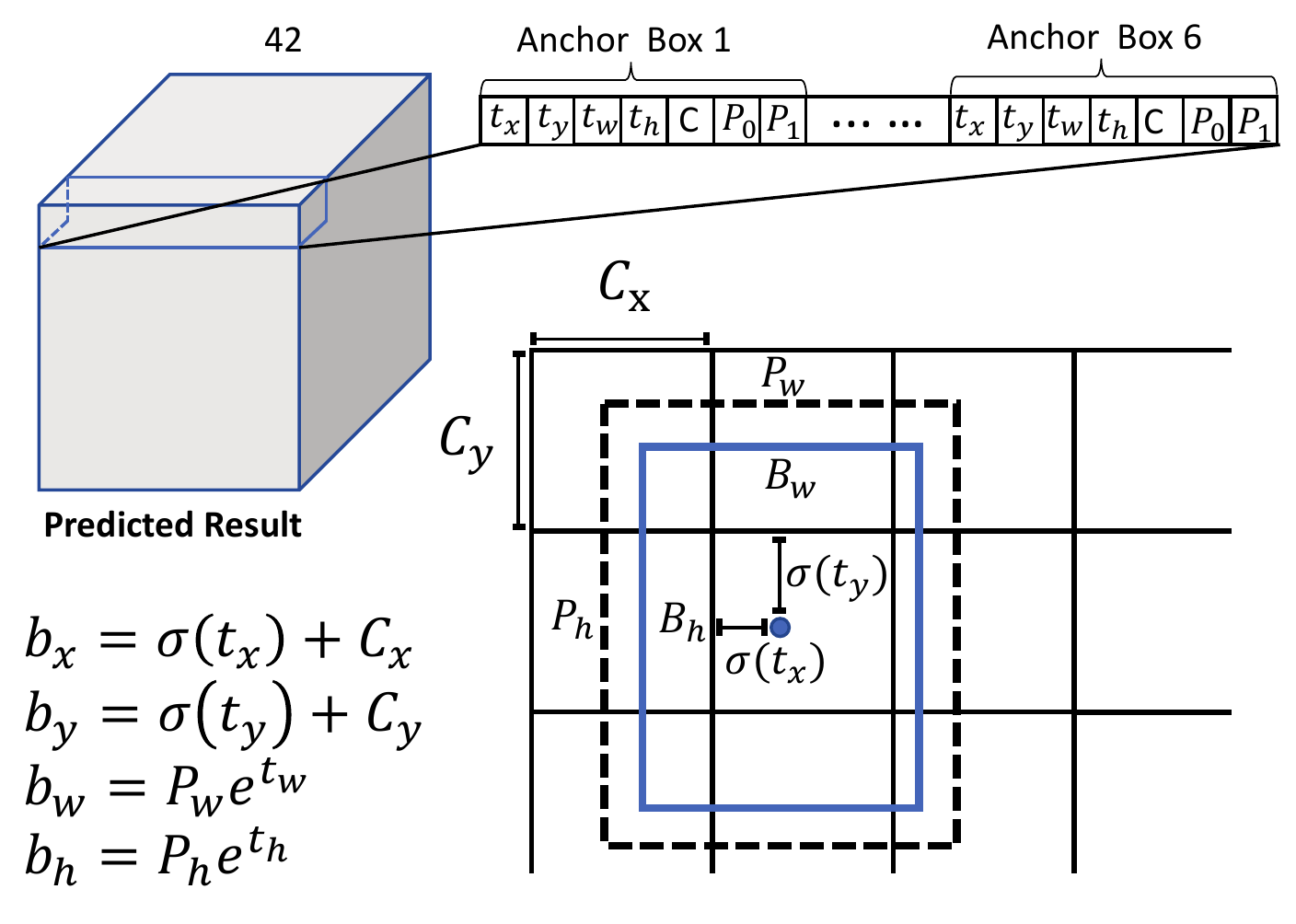}
\caption{The architecture of TOD-CNN head. Calculate the coordinates of the prediction 
box using the network output result and priori boxes. $t_x$, $t_y$, $t_w$ and $t_h$ 
are predicted by TOD-CNN for locating the bounding box. $P_0$ and $P_1$ represent the 
probability of sperm and impurity in the bounding box, respectively. $C$ is the 
confidence to determine whether there is an object in the bounding box.}
\label{fig6}
\end{figure}

In the head of TOD-CNN, 6 bounding boxes are predicted for each cell in the output 
feature map. For each bounding box, 7 coordinates ($t_x$, $t_y$, $t_w$, $t_h$, $C$, $P_0$ 
and $P_1$) are predicted, so the dimension of the predicted result in Fig.~\ref{fig6} is 42. For each cell, the offset from the upper left corner of the 
image is assumed to be ($C_x$, $C_y$), and the width and height of the corresponding 
a priori box are $P_w$ and $P_h$. The calculation method of center coordinates 
($b_x$ and $b_y$), width ($b_w$) and height ($b_h$) of predicted box is shown in 
Fig.~\ref{fig6}. Multi-label classification is applied to predict the categories  in each bounding box. Furthermore, due to the dense prediction method is 
applied to the head of TOD-CNN, the non-maximum value suppression method based on 
distance intersection over union~\cite{zheng2020DLFA} is used to remove bounding 
boxes with high overlap in the output results of the network.

\section{Experiments}
\subsection{Experimental Settings}
\subsubsection{Data Set}
A sperm microscopic video data set is released in our previous work~\cite{Chen-2021-SDAND} 
and it is used for the experiments of this paper. These sperm microscopic 
videos in the data set are obtained by a WLJY-9000 computer-aided sperm analysis 
system~\cite{hu2013COTS} under a $20 \times$ objective lens and a $20 \times$ 
electronic eyepiece. More than 278,000 objects are annotated in the data set: normal, 
needle-shaped, amorphous, cone-shaped, round, or multi-nucleated head sperms and 
impurities (such as bacteria, protein clumps, and bubbles). The object sizes range 
from approximately 5 to 50 $\mu$m$^2$. These objects are annotated by 14 reproductive 
doctors and biomedical scientists and verified by 6 reproductive doctors and biomedical 
scientists.

\begin{figure*}[htbp!]
\centering
\includegraphics[width=0.95\linewidth]{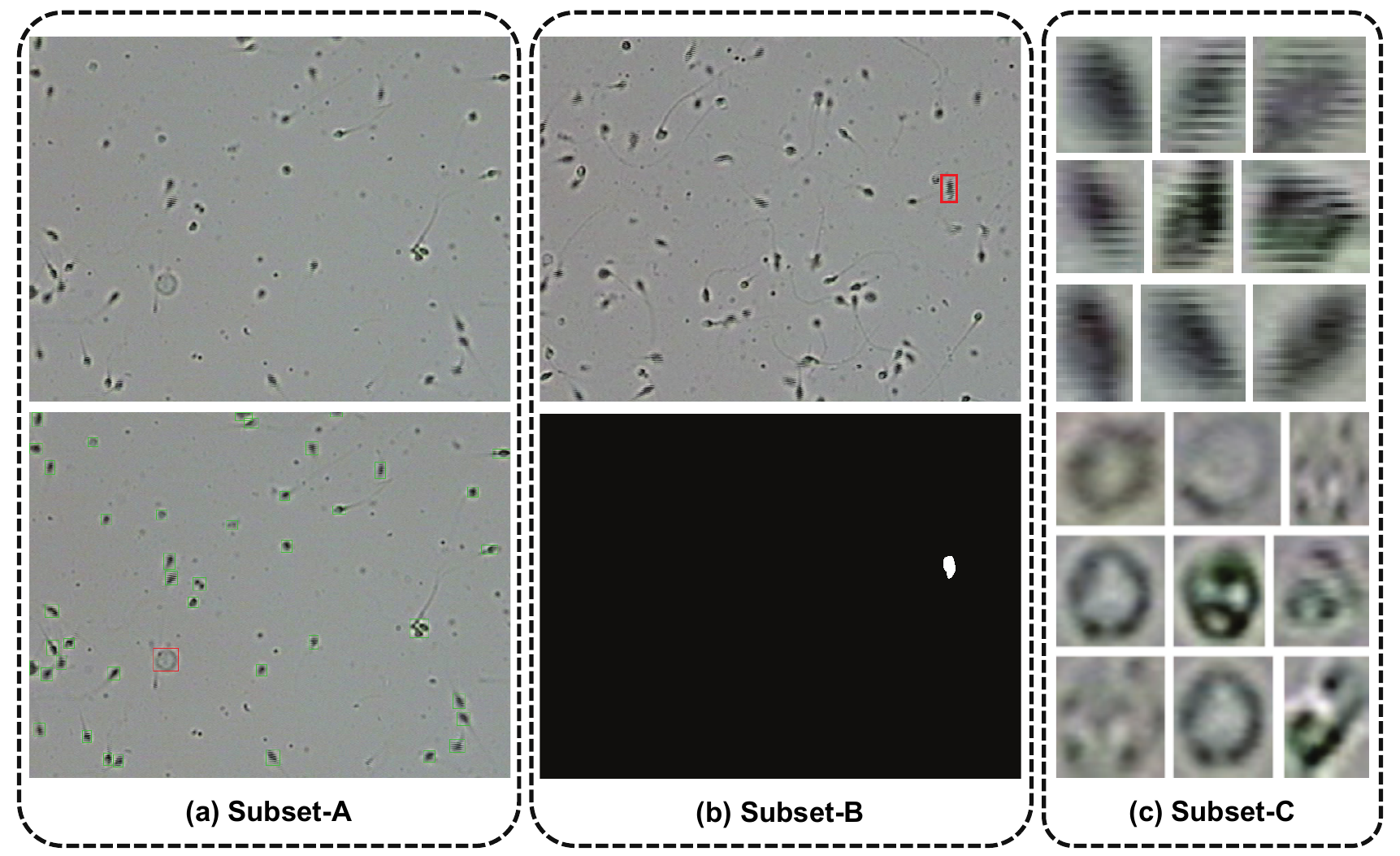}
\caption{An example of the sperm video data set. (a) The first row shows a frame in a sperm microscopic video and the bottom row is the corresponding annotation for object detection tasks. Sperms are in green boxes and impurities are in red boxes. 
(b) The first row shows a frame in sperm microscopic video and the bottom row shows the corresponding ground truth for object tracking tasks. 
(c) The first row shows individual sperm images and the bottom row shows individual impurity images for classification tasks.}
\label{fig7}
\end{figure*}

\begin{figure*}[htbp!]

	\centering
	\subfigure[Normal sperms]{
		\label{level.sub.1}
		\includegraphics[scale=1.65]{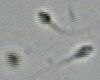}}
	\quad
	\quad
	\subfigure[Abnormal sperms]{
		\label{level.sub.2}
		\includegraphics[scale=1.65]{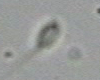}}

	\subfigure[Uncertain morphology (two-headed)]{
		\label{level.sub.3}
		\includegraphics[scale=1.7]{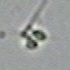}}
	\subfigure[Low contrast]{
		\label{level.sub.4}
		\includegraphics[scale=1.7]{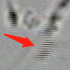}}
	\subfigure[Similar impurity]{
		\label{level.sub.5}
		\includegraphics[scale=1.44]{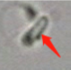}}
	\caption{Challenging cases for the detection of sperms. The positions pointed by the red arrow are the blur of sperm imaging caused by sperm movement and the impurity similar to sperm.}
		\label{fig1}
\end{figure*}

From 2017 to 2020, the collection and preparation of this data set took four years, 
including more than 278,000 annotated objects, as shown in Fig.~\ref{fig7}. Furthermore, the data set contains some hard-to-detect objects, such as uncertain morphology sperm, low contrast sperm, and similar impurities (as show in Fig.~\ref{fig1}), which greatly increases the difficulty of tiny object detection.

In this data set, Subset-A provides more than 125,000 objects with bounding box annotation 
and category information in 101 videos for tiny object detection task; Subset-B 
segments more than 26,000 sperms in 10 videos as ground truth for tiny object tracking 
task; Subset-C provides more than 125,000 independent images of sperms and impurities 
for tiny object classification task. Although Subset-C is not used in this work, it 
is still openly available to non-commercial scientific work.

\subsubsection{Training, Validation, and Test Data Setting}
We randomly divide the sperm microscopic video into training, validation, and test 
data sets at a ratio of 6:2:2. Therefore, we have 80 sperm microscopic videos and corresponding annotation information for training, and validation. The training set includes 2125 sperm microscopic images (77522 sperms and 2759 impurities), and validation set includes 668 sperm microscopic images (23173 sperms and 490 impurities).
And we have 21 sperm microscopic videos for testing, the test set includes 829 sperm microscopic images (20706 sperms and 1230 impurities).

\subsubsection{Experimental Environment}
The experiment is conducted by Python 3.7.0 in Windows 10 operating system. The models 
we use in this paper are implemented by Keras 2.1.5 framework with Tensorflow 1.13.1  
as the backend. Our experiment uses a workstation with Intel$^{(R)}$ Core$^{(TM)}$ i7-9700 CPU 
with 3.00GHz, 32GB RAM, and NVIDIA GEFORCE RTX 2080 8GB.

\subsubsection{Hyper Parameters}
The purpose of object detection task is to find all objects of interest in the image. Therefore, 
this task can be regarded as a combination of positioning and classification tasks. 
Therefore, as the loss function of the network, we use the complete intersection over 
union~\cite{zheng2020DLFA} (CIoU) function (location loss function) and the binary 
cross-entropy function (confidence and classification loss function), and then minimize them by Adam optimizer. For other hyper parameters, when freezing part of the layer 
training and unfreezing all layers, the batch size is set to 16 and 4, the training 
is 50 and 100 epochs, and the learning rate is set to $1\times10^{-3}$ and 
$1\times10^{-4}$, respectively. Besides, the cosine annealing 
scheduler~\cite{loshchilov2016SSGD} is used to adjust the learning rate. Besides,  when the 
loss value no longer drops, the training is terminated early.

\begin{table*}[b]
\centering
\caption{The definitions of evaluation metrics, where TP, TN, FP and FN represent 
True Positive, True Negative, False Positive and False Negative, respectively; 
N denotes the number of detected objects.}
\label{tab1}
\renewcommand\arraystretch{1.6}
\begin{tabular}{cccc}
\hline
Metric            & Definition & Metric      & Definition \\ \hline
Average Precision & $\frac{\sum_{i=1}^N \{Precision(i)\times Recall(i)\}}{Number~of~Annotations}$   &  Recall            & $\frac{TP}{TP + FN}$        \\ \hline
F$_1$ Score & $2 \times \frac{Precision \times Recall}{Precision + Recall}$    & Precision   & $\frac{TP}{TP + FP}$ \\ \hline   
\end{tabular}
\end{table*}

\subsection{Evaluation Metrics}
In order to quantitatively compare the performance of various object detection methods, 
different metrics are used to evaluate the detection results. Recall (Rec), Precision (Pre), 
F$_1$ Score (F$_1$), and Average Precision (AP) which can be used to evaluate the detection results.

Rec measures how many objects present in the annotation information are correctly detected. 
However, we cannot judge the detection result from the perspective of Rec alone. Pre 
measures how many objects detected by the model exist in the annotation information. 
The F$_1$ is the harmonic average of model Pre and Rec, and is an metric used to measure 
model performance. AP is a metric, which is widely used to evaluate the performance of object 
detection models. It can be obtained by calculating the area under the curve of Pre and Rec. 
It can evaluate object detection models from two aspects: Pre and Rec. The definitions 
of these evaluation metrics are provided in Table~\ref{tab1}.

\begin{figure}[htbp!]
\centering
\includegraphics[width=0.95\linewidth]{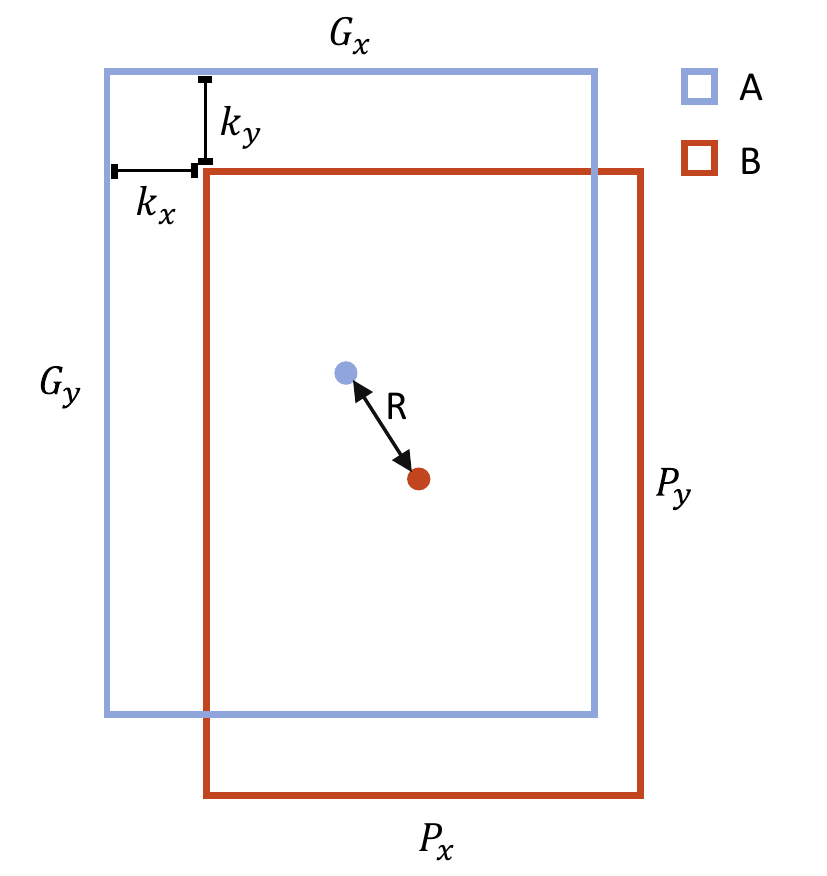}
\caption{The IoU calculation method.}
\label{fig8}
\end{figure}

The metrics in Table~\ref{tab1} are calculated based on True Positive, True Negative, 
False Positive, and False Negative. The intersection over union (IoU) is one of the 
evaluation criteria for evaluating whether the detected object is positive or negative. 
The calculation method of IoU is shown in Fig.~\ref{fig8} and Eq.~(\ref{eq1}).

\begin{align}
\label{eq1}
	IoU &= \frac{A \bigcap B}{A \bigcup B}\notag \\
	&= \frac{(G_x - k_x)(G_y - k_y)}{G_xG_y + P_xP_y - (G_x - k_x)(G_y - k_y)}
\end{align}

From Fig.~\ref{fig8} and Eq.~(\ref{eq1}), it can be found that the smaller the values 
of $G_x$, $G_y$, $P_x$, and $P_y$, the more sensitive the value of IoU to the changes 
of $k_x$ and $k_y$. The above phenomenon further illustrates that it is very difficult 
to detect tiny objects and it is unfair to use IoU alone to evaluate tiny object 
detection. Therefore, without affecting the sperm positioning, we propose a more 
suitable evaluation index. This indicator is a positive sample when the detected 
object meets two conditions at the same time: the first is that the detected object 
category is correct, and the second is that the IoU of the detection box and the ground 
truth box exceeds $B1$, or the IoU of the detection box and the ground truth box exceeds 
$B2$, and the distance between the center points of the two box does not exceed $R$ pixels.

\subsection{Evaluation of Sperm Detection Methods}
In order to prove the effectiveness of the proposed TOD-CNN method for sperm detection 
in sperm microscopic videos, we compared its detection results with other state-of-the-art methods, such as YOLO-v3~\cite{redmon2018yolov3}, YOLO-v4~\cite{bochkovskiy2020yolov4}, 
SSD~\cite{liu2016ssd}, RetinaNet~\cite{lin2017FLFD}, and Faster R-CNN~\cite{ren2016FRTR}. 
In the experiment process, each metric is calculated under the condition of $B1=0.5$, $B2=0.45$, and $R=3$. $B1$ and $B2$ represent the IoU value, and $R$ represents the pixel distance between the center of the predicted box and the center of the ground-truth box. Among them, whether $B1$ is greater than $0.5$ is a more common standard for evaluating positive and negative samples in the field of object detection~\cite{liu2020deep}. In addition, after our extensive experimental verification, the sperm object center coordinates obtained when $B2\geq0.45$ and $R\leq3$ have little effect on the sperm tracking task. Therefore, this paper adopts this standard to evaluate the experimental results.

\subsubsection{Compare with Other Methods}
In this part, we make a comparison between TOD-CNN and some state-of-the-art methods 
in terms of memory costs, training time, FPS, and detection 
performance.

\paragraph{\textbf{Evaluation of Memory, Time Costs and FPS}}To compare the memory 
costs, training time and FPS among TOD-CNN, YOLO-v4, YOLO-v3, SSD, RetinaNet and 
Faster R-CNN, we provide the details in Table~\ref{tab2}.

From Table~\ref{tab2}, we can find that the memory cost of TOD-CNN is 164 MB, the 
training time of TOD-CNN is around 119 min for 60 sperm microscopy videos, and the 
FPS is 34.7. In contrast, the memory cost and FPS of TOD-CNN are not optimal, but it 
considers both the model size and real-time performance. By comparing with YOLO-v3 and 
YOLO-v4, TOD-CNN has the minor memory cost. By comparing with RetinaNet and Faster R-CNN, 
TOD-CNN has faster detection speed. By comparing with SSD, TOD-CNN does not have better 
memory cost and real-time performance, but sperm detection ability of TOD-CNN is much 
better than SSD, which will be explained in detail in the next paragraph.

\begin{table}[htbp!]
\setlength\tabcolsep{2pt}
\centering
\caption{The memory costs, training time and FPS of TOD-CNN, YOLO-v4, YOLO-v3, SSD, 
RetinaNet and Faster R-CNN.}
\label{tab2}
\begin{tabular}{cccc}
\hline
Model       &  Memory Cost       & Training Time      & FPS          \\ \hline
TOD-CNN     &  164 MB            & \textbf{119 min}   & 35.7         \\ \hline
YOLO-v4     &  244 MB            & 135 min            & 28.4         \\ \hline
YOLO-v3     &  235 MB            & 374 min            & \textbf{37.0}\\ \hline
SSD         &  \textbf{91.2 MB}  & 280 min            & 31.5         \\ \hline
RetinaNet   &  139 MB            & 503 min            & 21.0         \\ \hline
Faster R-CNN&  108 MB            & 2753 min           & 7.8          \\ \hline
\end{tabular}
\end{table}

\begin{figure*}[htbp!]
	\centering
	\includegraphics[width=0.9 \linewidth]{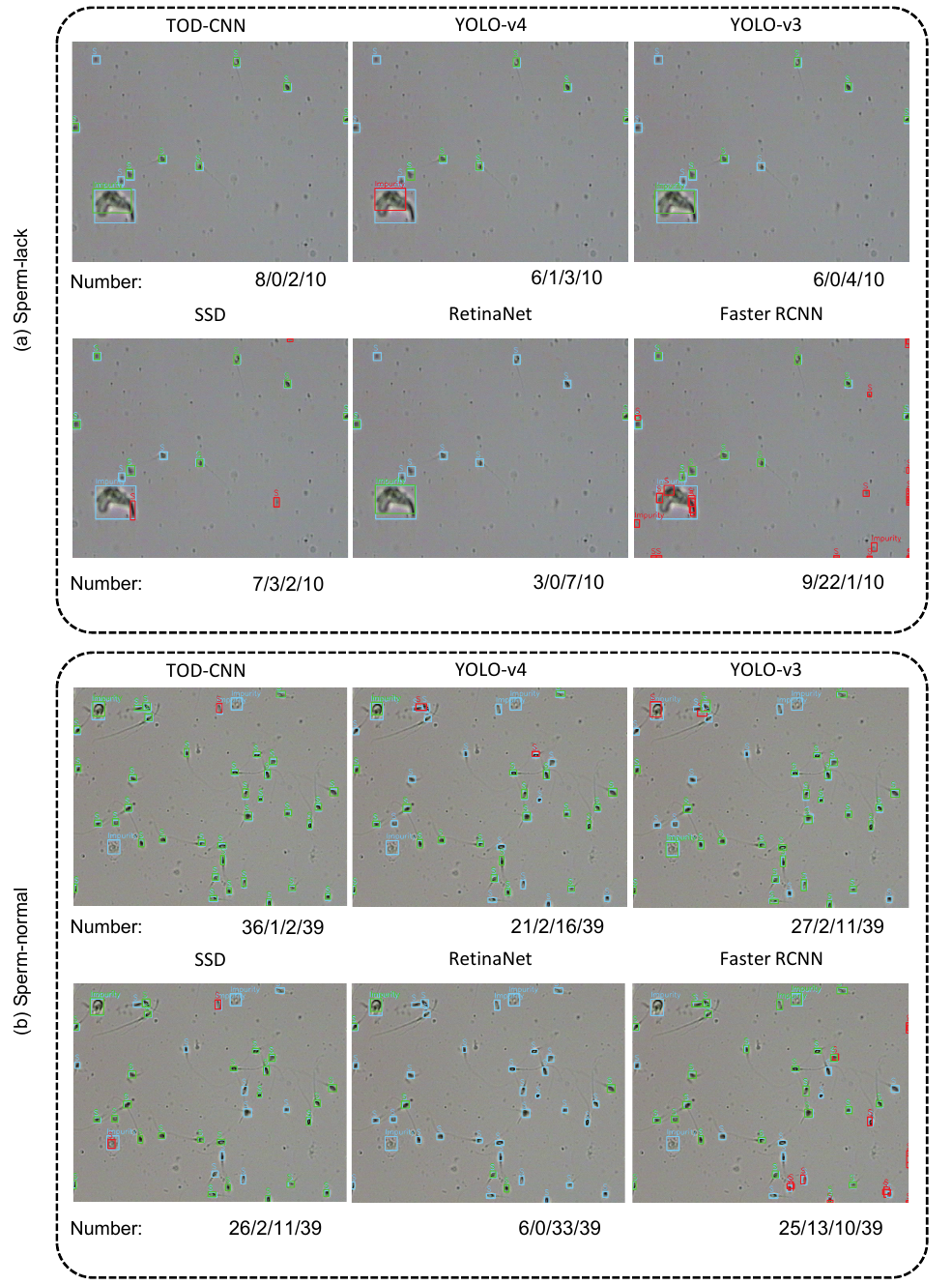}
	\caption{Comparison of TOD-CNN with YOLO-v4~\cite{bochkovskiy2020yolov4}, 
	YOLO-v3~\cite{redmon2018yolov3}, SSD~\cite{liu2016ssd}, RetinaNet~\cite{lin2017FLFD}, 
	and Faster RCNN~\cite{ren2015faster}. In these images, the blue boxes represent 
	the corresponding ground-truth, the green boxes correspond to the correctly 
	detected objects, and the red boxes correspond to the incorrectly detected objects. 
	The values represent the number of correctly detected objects/the number of 
	incorrectly detected objects/the number of objects in the annotation information 
	but are not detected/and the total number of objects in the annotation information.}
		\label{fig9}
\end{figure*}

\paragraph{\textbf{Evaluation of Sperm Detection Performance}}TOD-CNN is compared 
with existing object detection models using our data set. In Table~\ref{tab3}, we list the comparison with the best performance results of various models (SSD, 
RetinaNet, RetinaNet, and Faster R-CNN). In TOD-CNN, AP is nearly $20\%$ higher, 
F1 is nearly $12\%$ higher and Rec is nearly $22\%$ higher. 
Our Pre is about $6\%$ lower than the best performing model (RetinaNet). 
It is observed that our Rec is $75\%$ higher than that of RetinaNet, which shows that the number of detected objects obtained by TOD-CNN far exceeds RetinaNet. 
Overall, TOD-CNN outperforms existing models in sperm detection. Furthermore, a visual comparison of the models discussed above is shown in Fig.~\ref{fig9}.

\begin{table}[htbp!]
\centering
\caption{A comparison of detection results between TOD-CNN and existing models. (In [\%].)}
\label{tab3} 
\begin{tabular}{cccccc}
\hline
Models     & AP                 & F1             & Pre             & Rec           \\ \hline
TOD-CNN      & \textbf{85.60}   & \textbf{90.00} & 89.47           & \textbf{90.54}  \\
YOLO-v4     & 51.00             & 70.16          & 85.19           & 59.64       \\
YOLO-v3     & 42.93             & 64.36          & 78.36           & 54.60       \\
SSD         & 65.00             & 78.51          & 93.48           & 67.67       \\
RetinaNet   & 15.05             & 27.00          & \textbf{95.62}  & 15.72 \\
Faster RCNN & 35.76             & 55.28          & 46.57           & 67.99       \\ \hline 
 
\end{tabular}
\end{table}

From Fig.~\ref{fig9}, we can see that the correct detection case of TOD-CNN is only 
fewer than Faster RCNN in ``sperm-lack'' scenes (oligospermia), 
but our Pre is much higher than Faster RCNN~\cite{ren2015faster}. In ``sperm-normal'' 
scenes (healthy), the correct detection case of TOD-CNN is the best and our Pre and 
Rec are higher. 
By observing Fig.~\ref{fig9}, it is easy to understand why TOD-CNN has slightly lower 
Pre than SSD and RetinaNet, but other metrics are better than other models (the number 
of correct detections of TOD-CNN far exceeds other models).

Furthermore, to test the robustness of TOD-CNN against impurities in the microscopic 
videos, we have added 4,479 impurities into the experiments. 
The experimental results are shown in Table~\ref{tab4}, where TOD-CNN shows the best 
robustness against the effect of impurities compared to other models.

\begin{table}[h]
\setlength\tabcolsep{1pt}
\centering
	\caption{A comparison between TOD-CNN and existing models in the scene with impurity, 
where AP\_S, AP\_I and mAP represents AP of sperm, AP of impurity and mean AP, 
respectively. (In [\%].)}
\begin{tabular}{ccccccc}
\hline
Models     & AP\_S           & AP\_I          & mAP            & F1      & Pre      & Rec\\ \hline
TOD-CNN     & \textbf{85.60} & \textbf{57.33} & \textbf{71.47} & \textbf{88.57}   & 88.41    & \textbf{88.74}\\
YOLO-v4     & 51.00          & 30.00          & 40.50          & 69.61   & 84.76    & 59.06\\
YOLO-v3     & 42.93          & 35.90          & 39.42          & 63.80   & 78.34    & 53.81\\
Faster RCNN & 35.76          & 25.80          & 30.78          & 54.52   & 46.06    & 66.78\\
SSD         & 65.00          & 18.95          & 41.98          & 76.59   & 92.23    & 65.44\\
RetinaNet   & 15.05          & 33.84          & 24.44          & 28.51   & \textbf{95.36}    & 16.76\\ \hline
\label{tab4}
\end{tabular}
\end{table}

\subsubsection{Cross-validation Experiment}
To verify the reliability, stability and repeatability of TOD-CNN, we have performed 
five-fold cross-validation. The experimental results are shown in Table~\ref{Tab3}, 
where the mean values ($\mu$) of the four evaluation metrics is higher than $89\%$ 
except for AP, and AP is higher than $86\%$. It can be seen that TOD-CNN has good 
performance and repeatability. The standard deviation (STDEV) of F1 is $1.02\%$, the 
STDEV of two of the four evaluation metrics are below $1.40\%$, and only the STDEV of 
Pre is slightly higher ($2.05\%$), showing that TOD-CNN is relatively stable and reliable.

\begin{table}[h]
\centering
\caption{The detection results, $\mu$ and STDEV of the five-fold cross-validation 
experiments. (In [\%].)}
\begin{tabular}{cccccc}
\hline
Metrics & AP    & F1      & Pre   & Rec         \\ \hline
1       & 85.60 & 90.00   & 89.47 & 90.54   \\
2       & 84.90 & 90.11   & 92.76 & 87.61   \\
3       & 88.80 & 92.78   & 95.19 & 90.48   \\
4       & 86.37 & 91.33   & 94.12 & 88.66   \\
5       & 87.29 & 90.65   & 91.15 & 90.16   \\
$\mu$   & 86.59 & 90.97   & 92.54 & 89.49   \\
STDEV   & 1.36  & 1.02    & 2.05  & 1.16    \\ \hline
\label{Tab3}
\end{tabular}
\end{table}

\subsubsection{Sperm Tracking}
\begin{figure*}[h]
	\centering
	\includegraphics[width=0.8\linewidth]{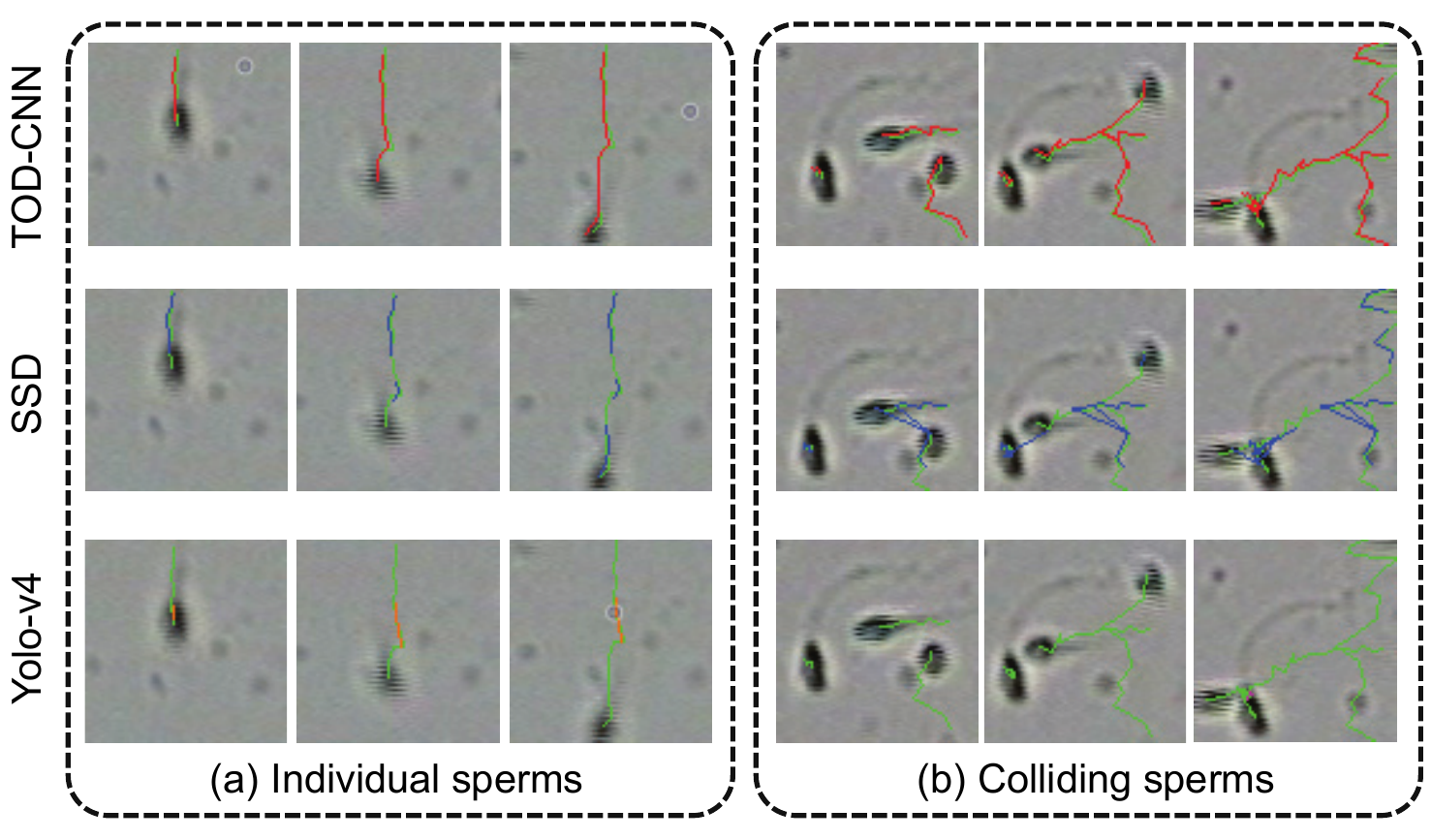}
	\caption{An example of sperm tracking results. Green lines represent the 
	actual trajectories based ground truth; red, blue and orange lines denote the 
	tracking trajectories of TOD-CNN, SSD and Yolo-v4, respectively.}
		\label{fig10}
\end{figure*}

The ultimate goal of sperm detection is to find the sperm trajectories and calculate the relevant parameters for clinical diagnosis. 
TOD-CNN and two models with better detection results (YOLO-v4 and SSD) are are compared for sperm tracking in Table 3. 
Based on the detection result of each model, we use the $k$NN algorithm to match 
sperms in adjacent video frames to the actual trajectories marked in Subset-B. 
The visualization results are shown in Fig.~\ref{fig10}. 
We can observe that our tracking trajectories are very close to the actual ones, 
and the trajectory discontinuity or incorrect tracking is rarely occurred due to the 
stronger detection capability of TOD-CNN.

In addition, we calculate three important motility parameters of sperms on Subset-B, 
including the Straight Line Velocity (VSL), Curvilinear Velocity (VCL) and Average 
Path Velocity (VAP)~\cite{o2002TEOC} of actual trajectories, with TOD-CNN, SSD and 
Yolo-v4, respectively. Comparing with the actual trajectories, the error rates of VSL, 
VCL and VAP calculated with TOD-CNN (10.15\%, 5.09\% and 8.95\%) are significantly 
lower than that of SSD (41.58\%, 5.01\% and 17.40\%) and Yolo-v4 (12.73\%, 36.12\% 
and 19.65\%). Based on  VCL, VSL, and VAP, an experienced threshold value from a 
clinical doctor is set to determine whether a sperm is motile to calculate the 
corresponding progressive motility (PR). The error between PR obtained by TOD-CNN 
tracking results and  doctors' diagnosis results are all within 9\%. The experimental 
result shows that our TOD-CNN can assist doctors in clinical work.

\subsubsection{A Python-based Graphical User Interface}
\begin{figure}[htbp!]
	\centering
	\includegraphics[width=0.98 \linewidth]{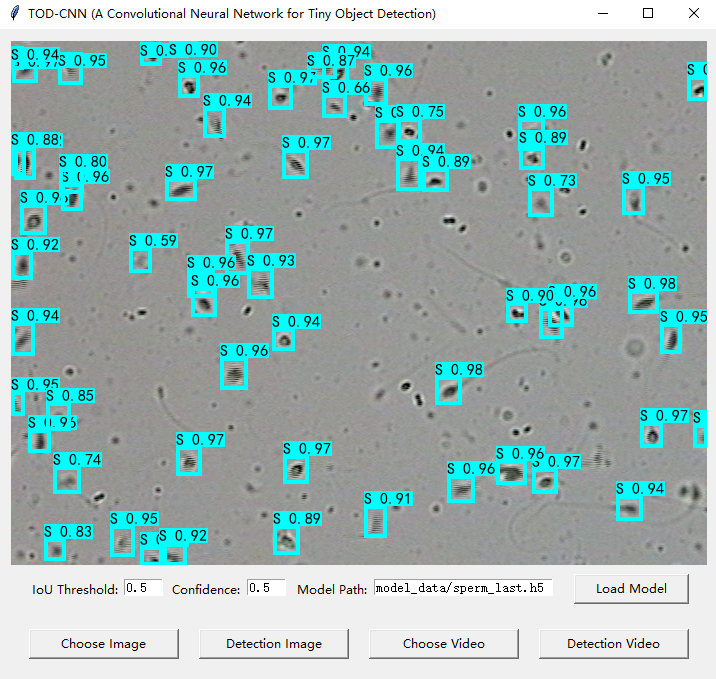}
	\caption{GUI of TOD-CNN for detecting sperms in microscopic videos or images.}
		\label{fig11}
\end{figure}

To conveniently use TOD-CNN to detect tiny objects in microscopic videos and images, 
we design a Python-based GUI (Fig.~\ref{fig11}) that can help users to control the 
Intersection of Union (IoU) threshold and confidence according to their own needs to
 achieve the desired test performance. 
Besides, users can load Model Path to use their own setting/weights for tiny object 
detection. This GUI is compatible with videos (such as ``.mp4'' and ``.avi'') and 
images (such as ``.png'' and ``.jpg'') in various formats.

\section{Conclusion and Discussion}
\begin{figure}[htbp]
	\centering
	\subfigure[]{
		\label{fig12.sub.1}
		\includegraphics[scale=1.5]{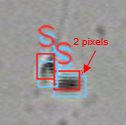}}
	\subfigure[]{
		\label{fig12.sub.2}
		\includegraphics[scale=1.5]{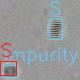}}
		
	\subfigure[]{
		\label{fig12.sub.3}
		\includegraphics[scale=1.5]{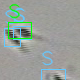}}
	\subfigure[]{
		\label{fig12.sub.4}
		\includegraphics[scale=1.5]{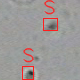}}
	\caption{Visualized results of some typical detection failures of TOD-CNN. 
	The red and green boxes represent the detection results, the blue boxes 
	represent the ground truth, S represents sperms and Impurity represents impurities.}
		\label{fig12}
\end{figure}

We develop and present a public, massive and high-quality data set for sperm detection, 
tracking and classification, and this data set now is published and available online. 
We also provide a one-stage CNN model (TOD-CNN) on Subset-A for tiny object detection 
in real-time, which can accurately detect sperms in videos and images. 
However, TOD-CNN fails in some cases and cannot detect sperms completely or accurately. 
The example of incorrect detection results are shown in Fig.~\ref{fig12}.

In Fig.~\ref{fig12.sub.1}, we can see that the detection boxes can surround sperms 
correctly. However, due to the small size of the ground truth boxes, a minor position 
offset (one or two pixels) causes the IoU between detection and ground truth boxes 
to be lower than 0.5. 
In Fig.~\ref{fig12.sub.2}, due to the movement of the sperms, the thickness of the 
semen wet film and noticeable interference fringes in sperm videos, it may lose valuable 
information and lead to errors in detection. Also, because some impurities have very close visual information to sperms, TOD-CNN incorrectly detects the impurities as sperms. 
In Fig.~\ref{fig12.sub.3}, for the sperms appearing on figure edges, it is difficult 
to explore the complete information and sometimes these sperms are missed in detection. 
To ensure the annotation information reliability, when we marked sperms in videos, we 
only choose sperms without controversy. 
In Fig.~\ref{fig12.sub.4}, the detected sperms may be located deeply in the semen wet film. Because of its unclear imaging, it is difficult to distinguish whether it is a sperm or an impurity and it is not annotated in our data set.

In future work, we will continue to integrate related optimization algorithms to improve the performance of TOD-CNN, such as monarch butterfly optimization~\cite{wang2019monarch}, 
earthworm optimization algorithm~\cite{wang2018earthworm}, 
elephant herding optimization~\cite{wang2015elephant, li2020elephant}, 
moth search algorithm~\cite{wang2018moth}, 
slime mould algorithm~\cite{li2020slime}, 
hunger games search~\cite{yang2021hunger}, 
Runge Kutta optimizer~\cite{li2021survey}, 
colony predation algorithm~\cite{tu2021colony, li2022review}, and
Harris hawks optimization~\cite{heidari2019harris}.

\section*{Acknowledgements}
This work is supported by the ``National Natural Science Foundation of China'' 
(No. 61806047). 
We thank Miss Zixian Li and Mr. Guoxian Li for their important discussion. 

\section*{Declaration of Competing Interest} 
The authors declare that they have no conflict of interest.

\bibliographystyle{elsarticle-num}
\bibliography{Rf}

\end{document}